\documentclass{article} 
\usepackage{amsfonts}
\usepackage{amsmath}
\usepackage{iclr2017_workshop,times}
\usepackage{hyperref}
\usepackage{url}
\usepackage{cleveref}
\usepackage{graphicx}
\usepackage{enumitem}

\usepackage{color}
\usepackage[procnames]{listings}
\usepackage{textcomp}
\usepackage{setspace}

\definecolor{gray}{gray}{0.4}
\definecolor{green}{rgb}{0,0.5,0}
\definecolor{lightgreen}{rgb}{0,0.7,0}
\definecolor{purple}{rgb}{0.5,0,0.5}
\definecolor{darkred}{rgb}{0.5,0,0}
\lstnewenvironment{python}[1][]{
\lstset{
language=python,
basicstyle=\ttfamily\small\setstretch{1},
stringstyle=\color{green},
showstringspaces=false,
alsoletter={1234567890},
emph={access,and,as,break,class,continue,def,del,elif,else,%
except,exec,finally,for,from,global,if,import,in,is,%
lambda,not,or,pass,print,raise,return,try,while,assert},
emphstyle=\color{darkred}\bfseries,
emph={[2]self},
emphstyle=[2]\color{gray},
emph={[4]ArithmeticError,AssertionError,AttributeError,BaseException,%
DeprecationWarning,EOFError,Ellipsis,EnvironmentError,Exception,%
False,FloatingPointError,FutureWarning,GeneratorExit,IOError,%
ImportError,ImportWarning,IndentationError,IndexError,KeyError,%
KeyboardInterrupt,LookupError,MemoryError,NameError,None,%
NotImplemented,NotImplementedError,OSError,OverflowError,%
PendingDeprecationWarning,ReferenceError,RuntimeError,RuntimeWarning,%
StandardError,StopIteration,SyntaxError,SyntaxWarning,SystemError,%
SystemExit,TabError,True,TypeError,UnboundLocalError,UnicodeDecodeError,%
UnicodeEncodeError,UnicodeError,UnicodeTranslateError,UnicodeWarning,%
UserWarning,ValueError,Warning,ZeroDivisionError,abs,all,any,apply,%
basestring,bool,buffer,callable,chr,classmethod,cmp,coerce,compile,%
complex,copyright,credits,delattr,dict,dir,divmod,enumerate,eval,%
execfile,exit,file,filter,float,frozenset,getattr,globals,hasattr,%
hash,help,hex,id,input,int,intern,isinstance,issubclass,iter,len,%
license,list,locals,long,map,max,min,object,oct,open,ord,pow,property,%
quit,range,raw_input,reduce,reload,repr,reversed,round,set,setattr,%
slice,sorted,staticmethod,str,sum,super,tuple,type,unichr,unicode,%
vars,xrange,zip},
emphstyle=[4]\color{purple}\bfseries,
upquote=true,
morecomment=[s][\color{lightgreen}]{"""}{"""},
commentstyle=\color{gray}\slshape,
literate={>>>}{\textbf{\textcolor{darkred}{>{>}>}}}3%
         {...}{{\textcolor{gray}{...}}}3,
procnamekeys={def,class},
procnamestyle=\color{blue}\textbf,
}}{}

\title{Training a Subsampling Mechanism in \mbox{Expectation}}

\author{Colin Raffel \& Dieterich Lawson\thanks{Work done as members of the Google Brain Residency program} \\
Google Brain\\
\texttt{craffel@gmail.com, dieterichl@google.com}
}

\begin{document}

\maketitle

\begin{abstract}
We describe a mechanism for subsampling sequences and show how to compute its expected output so that it can be trained with standard backpropagation.  We test this approach on a simple toy problem and discuss its shortcomings.
\end{abstract}

\section{Subsampling Sequences}
\label{sec:definition}

Consider a mechanism which, given a sequence of vectors $\mathbf{s} = \{s_0, s_1, \ldots, s_{T - 1}\}, s_t \in \mathbb{R}^d$, produces a sequence of ``sampling probabilities'' $\mathbf{e} = \{e_0, e_1, \ldots, e_{T - 1}\}, e_t \in [0, 1]$ which denote the probability of including $s_t$ in the output sequence $\mathbf{y} = \{y_0, y_1, \ldots, y_{U-1}\}$.
Producing $\mathbf{y}$ from $\mathbf{s}$ and $\mathbf{e}$ is encapsulated by the following pseudo-code and visualized in \cref{fig:subsampling_schematic} (appendix):

\begin{python}
# Initialize y as an empty sequence
y = []
for t in {0, 1, ..., T - 1}:
    # Draw a random number in [0, 1] and compare to e[t]
    if rand() < e[t]:
         # Add s[t] to y with probability e[t]
         y.append(s[t])
\end{python}

We call this a ``subsampling mechanism'', because by construction, $U \le T$, and each element of $\mathbf{y}$ is drawn directly from $\mathbf{s}$.
The ability to subsample a sequence has various applications:
\begin{itemize}[topsep=0pt,itemsep=-1pt,leftmargin=20pt]
\item When the input sequence $\mathbf{s}$ is oversampled (i.e.\ each element $s_t$ contains much the same information as $s_{t - 1}$), subsampling can be an effective way of shortening the sequence without discarding useful information.  Using a shorter sequence can facilitate the use of recurrent network models, which have difficulties with long-term dependencies \citep{bengio1994learning,hochreiter1997long}.  Simple subsampling schemes such as choosing every other element of $\mathbf{s}$ have  proven effective in tasks such as speech recognition \citep{chan2015listen}.
\item The mechanism can be used in sequence transduction tasks where the output sequence is shorter than the input.  We contrast this approach with the commonly used Connectionist Temporal Classification loss \citep{graves2006connectionist} because subsampling actually shortens the sequence (instead of inserting blanks) and can be inserted arbitrarily into a neural network model (instead of specifically being a loss function).  It also implicitly produces a monotonic alignment between elements in $\mathbf{s}$ and $\mathbf{y}$; such alignments have proven to be useful \citep{bahdanau2014neural}.
\item Applying this subsampling operation multiple times could build a hierarchy of shorter and shorter sequences which capture structure at different scales.  A similar approach was recently shown to be effective in langauge modeling tasks \citep{chung2016hierarchical}.
\end{itemize}
Motivated by these applications, in this extended abstract we present a method for training this subsampling mechanism in expectation, i.e.\ without sampling.
We then test this approach on a simple toy problem and study the resulting model's behavior.
Finally, we discuss shortcomings of our approach and possibilities for future work.

\section{Training in Expectation}
\label{sec:expectation}

We are interested in including the mechanism defined in the previous section in the midst of a neural network model.
However, the sampling process used to construct $\mathbf{y}$ precludes the use of standard backpropagation.
A common approach to this issue is to optimize the model according to the expected (or mean-field) output \citep{graves2016adaptive,bahdanau2014neural}.
The following analysis shows how to employ this approach to our proposed subsampling mechanism using a dynamic program which analytically computes $p(y_m = s_n)$.

First, observe that $p(y_0 = s_0) = e_0$, i.e. the probability that the first output is the first entry in the sequence is just the probability of sampling at time 0.
Next, in order for $y_0 = s_1$, we need $y_0 \ne s_0$ so $p(y_0 = s_1)$ is the probability that $s_0$ was not sampled at time 0 and that $s_1$ was, giving $p(y_0 = s_1) = e_1(1 - e_0)$.
Continuing on in this way, we see that $p(y_0 = s_n) = e_n\prod_{i = 0}^{n - 1}(1 - e_i)$ or, in words, the probability that the first output element $y_0$ is a given element in the sequence $s_n$ is the probability that none of $s_0, \ldots, s_{n - 1}$ were sampled multiplied by the probability of sampling $s_n$.  Second, observe that $p(y_m = s_n) = 0$ when $n < m$ because in order for the output sequence to be of length $m$, at least $m - 1$ symbols must already have been sampled.
If $n < m$, this relation is violated.  Finally, in order for $y_m = s_n$ in general, we must have that $y_{m-1}=s_j \in {s_0, \ldots, s_{n-1}}$ (i.e. the previous output must be one of the states before $s_n$), none of $s_{j + 1}, \ldots, s_{n-1}$ may be sampled at time $m$, and $s_n$ is sampled at time $m$.
To compute $p(y_m = s_n)$, we need to sum over all of the the possible cases $y_{m-1} \in \{s_0, \ldots, s_{n-1}\}$.  The probability of a single case is the combined probability that $s_n$ is sampled, that $y_{m - 1} = s_j$, and that none of $s_{j + 1}, \ldots, s_{n-1}$ are sampled at time $m$.  We visualize these possibilities in \cref{fig:paths} (appendix).  Summing over the possible $j$ yields
\begin{equation}
p(y_m = s_n) = e_n \sum_{j = 0}^{n - 1}\left(p(y_{m - 1} = s_j) \prod_{i = j + 1}^{n - 1}(1 - e_i)\right)
\label{eq:p_y_m_eq_s_n}
\end{equation}
where for convenience we define the special case $\prod_{i = n}^m \bullet = 1$ when $n > m$.
Once we compute $p(y_m = s_n)$, it is straightforward to find the expected value of $y_m$ simply by computing $\sum_n s_n p(y_m = s_n)$.
Note $p(y_m = s_n) = e_n ((1 - e_{n - 1})p(y_m = s_{n - 1})/e_{n - 1} + p(y_{m - 1} = s_{n - 1}))$; it follows that each term $p(y_m = s_n)$ can be computed in $\mathcal{O}(1)$ time by reusing the already-computed terms $p(y_m = s_{n - 1})$ and $p(y_{m - 1} = s_{n - 1})$.
The resulting dynamic program allows all the terms $p(y_m = s_n)$ to be computed in $\mathcal{O}(T^2)$ time.

Note that $\sum_n p(y_m = s_n) \le 1$ depending on the values of $\mathbf{e}$, so these probabilities may not form a valid probability distribution.
Computing the expectation as-is without further normalization effectively associates any additional probability to an implicit zero vector in $\mathbb{R}^d$, which is the convention we will use for the remainder of this extended abstract.

\section{Toy Problem Experiment}
\label{sec:experiment}

To evaluate the feasibility of this approach, we tested it on the following toy problem:
Consider a length-$T$ sequence $\mathbf{x}$ of symbols $[0, 1, 2]$ which occur with equal probability.
The output is produced as follows for $t \in \{0, \ldots, T - 1\}$, beginning with an empty memory:
\begin{enumerate}[topsep=1pt,itemsep=-1pt,leftmargin=20pt]
\item If $x_t$ is 0, don't output anything and maintain the current memory state.
\item If $x_t$ is 1 or 2 and our memory is empty, place $x_t$ in memory and don't output anything.
\item If $x_t$ is 1 and we have 1 in our memory, output a 0 and empty the memory.
\item If $x_t$ is 2 and we have 2 in our memory, output a 0 and empty the memory.
\item If $x_t$ is 1 and we have 2 in our memory, output a 2 and empty the memory.
\item If $x_t$ is 2 and we have 1 in our memory, output a 1 and empty the memory.
\end{enumerate}
We also define special cases where if $T = 1$, the output is $x_0$; if $x_t = 0 \; \forall \; t \in \{0, \ldots, T - 1\}$, the output is 0; and if all entries of $x_t$ are 0 except one, the output is the single nonzero entry. An example input-output pair for this toy problem is shown in \cref{fig:toy_example} (appendix).

We utilized the following model:
\begin{align}
s_t &= \mathrm{LSTM}(x_t, s_{t - 1})\\
e_t &= \sigma(W_{he}^\top s_{t - 1} + W_{xe}^\top x_{t - 1} + b_e)\\
\label{eq:e_t}
y_t &= \mathrm{softmax}\left(W_y^\top \sum_{n = 0}^{T - 1} p(y_t = s_n)s_n + b_y\right)
\end{align}
where $x_t \in \mathbb{R}^3$ is the one-hot encoding of the input sequence, $\mathrm{LSTM}$ is a long short-term memory RNN \citep{hochreiter1997long} with state dimensionality 100, $W_{he} \in \mathbb{R}^{100 \times 1}, W_{xe} \in \mathbb{R}^{3 \times 1}, b_e \in \mathbb{R}$ are the weight matrices and bias scalar for computing emission probabilities, $\sigma(\cdot)$ is the logistic sigmoid function, and $W_y \in \mathbb{R}^{100 \times 3}, b_y \in \mathbb{R}^3$ are the weight matrix and bias vector of the output softmax function.
The $p(y_t = s_n)$ terms are computed as described in \cref{sec:expectation}.

We fed minibatches of 100 sequences of randomly chosen $[0, 1, 2]$ values, encoded as one-hot vectors, to the network.
The network was trained with categorical cross-entropy against analytically computed targets using Adam with the learning hyperparameters suggested in \citep{kingma2015adam}.
We computed the network's accuracy on a separately generated test set that it was not trained on.
As proposed in \citep{zaremba2014learning}, we found it beneficial to use a simple curriculum learning \citep{bengio2009curriculum} strategy where the loss was only computed for the first $T^\prime$ elements of the output sequence, where $T^\prime$ was uniformly sampled from the values $\{1, 2, \ldots, T\}$ for each minibatch.

\begin{figure}[h]
\begin{center}
\includegraphics[width=\textwidth]{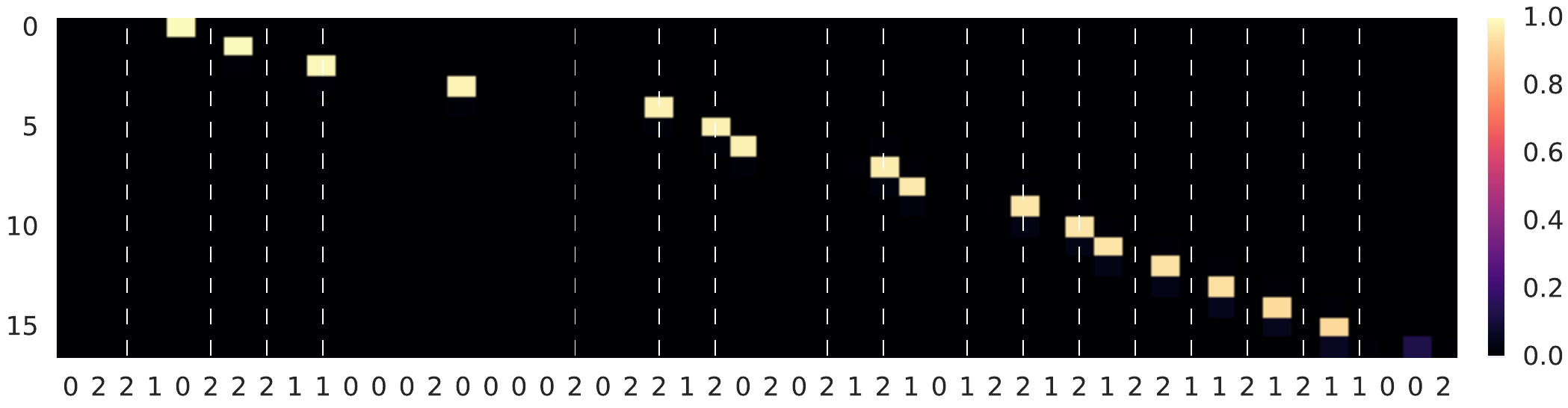}
\end{center}
\caption{$p(y_t = s_n)$ for an example test sequence of length $50$.  The sequence is shown on the x-axis, with dashed vertical lines denoting where we might expect the model to emit symbols.  The y-axis shows the output sequence index.  For reference, the correct output for this sequence is $\{0, 1, 0, 0, 0, 0, 1, 0, 1, 0, 0, 1, 1, 2, 1, 1, 0\}$.}
\label{fig:p_y_t_eq_s_n}
\end{figure}

For all values of $T$ we tried (up to $T = 500$), the network was able to achieve $> 98\%$ accuracy on the held-out test set after training for a modest number of minibatches (around 10,000).
To get a picture of the qualitative behavior of the model, we plot the matrix $p(y_t = s_n)$ for an example test sequence with $T = 50$ in \cref{fig:p_y_t_eq_s_n}.
Note that emissions do not occur exactly when the model has seen sufficient input to produce them, i.e.\ once it sees a second nonzero input.
In this particular case, this caused the model to emit one too few symbols.
To facilitate further research, we provide a TensorFlow implementation of our approach.\footnote{\url{https://github.com/craffel/subsampling_in_expectation}}

While we have shown that our model can quickly learn the desired behavior on a toy problem, we had issues applying this approach to real-world problems, which we attribute primarily to two factors:
First, while a stated goal of the subsampling mechanism is to produce shorter sequences, the $\mathcal{O}(T^2)$ complexity of computing the terms $p(y_m = s_n)$ precludes its practical use on problems with large $T$.
Second, the use of a sigmoid in \cref{eq:e_t} and the cumulative product in \cref{eq:p_y_m_eq_s_n} can result in vanishing gradients in practice.
The first issue could be mitigated by greedy approximations to the procedure outlined in \cref{sec:expectation}, for example by selecting which items in $s_t$ are chosen using discrete latent variables and training with reinforcement learning methods as has been done in recent work \citep{luo2016learning}.
We hope the encouraging results and analysis presented here inspires future work on utilizing learnable subsampling mechanisms in neural networks.

\bibliography{iclr2017_workshop}

\begin{thebibliography}{11}
\providecommand{\natexlab}[1]{#1}
\providecommand{\url}[1]{\texttt{#1}}
\expandafter\ifx\csname urlstyle\endcsname\relax
  \providecommand{\doi}[1]{doi: #1}\else
  \providecommand{\doi}{doi: \begingroup \urlstyle{rm}\Url}\fi

\bibitem[Bahdanau et~al.(2014)Bahdanau, Cho, and Bengio]{bahdanau2014neural}
Dzmitry Bahdanau, Kyunghyun Cho, and Yoshua Bengio.
\newblock Neural machine translation by jointly learning to align and
  translate.
\newblock \emph{arXiv preprint arXiv:1409.0473}, 2014.

\bibitem[Bengio et~al.(1994)Bengio, Simard, and Frasconi]{bengio1994learning}
Yoshua Bengio, Patrice Simard, and Paolo Frasconi.
\newblock Learning long-term dependencies with gradient descent is difficult.
\newblock \emph{IEEE Transactions on Neural Networks}, 5\penalty0 (2):\penalty0
  157--166, 1994.

\bibitem[Bengio et~al.(2009)Bengio, Louradour, Collobert, and
  Weston]{bengio2009curriculum}
Yoshua Bengio, J{\'e}r{\^o}me Louradour, Ronan Collobert, and Jason Weston.
\newblock Curriculum learning.
\newblock In \emph{Proceedings of the 26th International Conference on Machine
  Learning}, pp.\  41--48, 2009.

\bibitem[Chan et~al.(2015)Chan, Jaitly, Le, and Vinyals]{chan2015listen}
William Chan, Navdeep Jaitly, Quoc~V. Le, and Oriol Vinyals.
\newblock Listen, attend and spell.
\newblock \emph{arXiv preprint arXiv:1508.01211}, 2015.

\bibitem[Chung et~al.(2016)Chung, Ahn, and Bengio]{chung2016hierarchical}
Junyoung Chung, Sungjin Ahn, and Yoshua Bengio.
\newblock Hierarchical multiscale recurrent neural networks.
\newblock \emph{arXiv preprint arXiv:1609.01704}, 2016.

\bibitem[Graves(2016)]{graves2016adaptive}
Alex Graves.
\newblock Adaptive computation time for recurrent neural networks.
\newblock \emph{arXiv preprint arXiv:1603.08983}, 2016.

\bibitem[Graves et~al.(2006)Graves, Fern{\'a}ndez, Gomez, and
  Schmidhuber]{graves2006connectionist}
Alex Graves, Santiago Fern{\'a}ndez, Faustino Gomez, and J{\"u}rgen
  Schmidhuber.
\newblock Connectionist temporal classification: Labelling unsegmented sequence
  data with recurrent neural networks.
\newblock In \emph{Proceedings of the 23rd International Conference on Machine
  learning}, pp.\  369--376, 2006.

\bibitem[Hochreiter \& Schmidhuber(1997)Hochreiter and
  Schmidhuber]{hochreiter1997long}
Sepp Hochreiter and J{\"u}rgen Schmidhuber.
\newblock Long short-term memory.
\newblock \emph{Neural Computation}, 9\penalty0 (8):\penalty0 1735--1780, 1997.

\bibitem[Kingma \& Ba(2015)Kingma and Ba]{kingma2015adam}
Diederik~P. Kingma and Jimmy Ba.
\newblock Adam: A method for stochastic optimization.
\newblock In \emph{Proceedings of the 3rd International Conference on Learning
  Representations}, 2015.

\bibitem[Luo et~al.(2016)Luo, Chiu, Jaitly, and Sutskever]{luo2016learning}
Yuping Luo, Chung-Cheng Chiu, Navdeep Jaitly, and Ilya Sutskever.
\newblock Learning online alignments with continuous rewards policy gradient.
\newblock \emph{arXiv preprint arXiv:1608.01281}, 2016.

\bibitem[Zaremba \& Sutskever(2014)Zaremba and Sutskever]{zaremba2014learning}
Wojciech Zaremba and Ilya Sutskever.
\newblock Learning to execute.
\newblock \emph{arXiv preprint arXiv:1410.4615}, 2014.

\end{thebibliography}
\bibliographystyle{iclr2017_workshop}

\clearpage

\appendix

\section{Figures}

In this appendix we provide additional figures to help illustrate some of the concepts presented in this extended abstract.

\begin{figure}[h]
\begin{center}
\includegraphics[width=\textwidth]{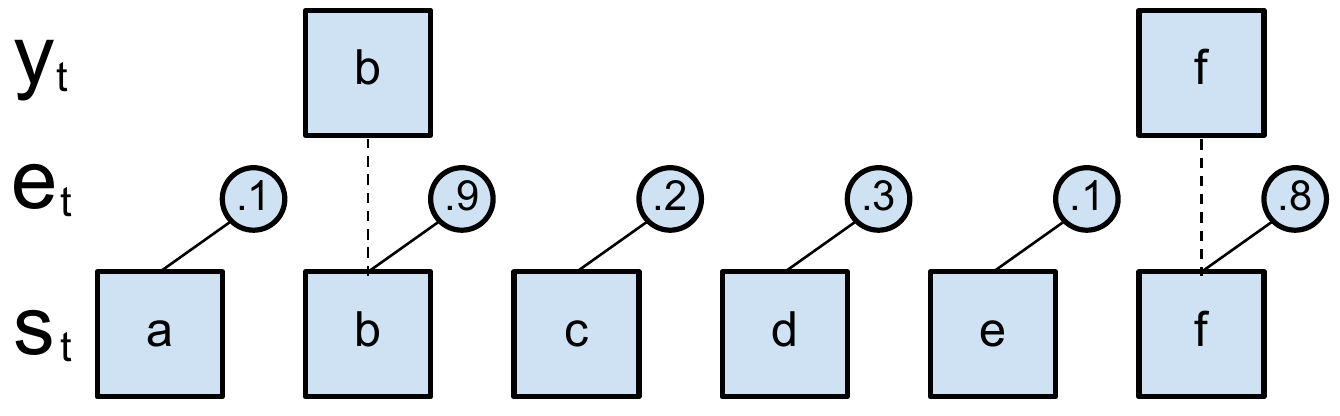}
\end{center}
\caption{Illustration of the subsampling process described in \cref{sec:definition}.  Each element $s_t$ of $\mathbf{s}$ is included in the output sequence $\mathbf{y}$ with probability $e_t$.  In this case, the second and final elements of $\mathbf{s}$ were sampled.}
\label{fig:subsampling_schematic}
\end{figure}

\begin{figure}[h]
\begin{center}
\includegraphics[width=\textwidth]{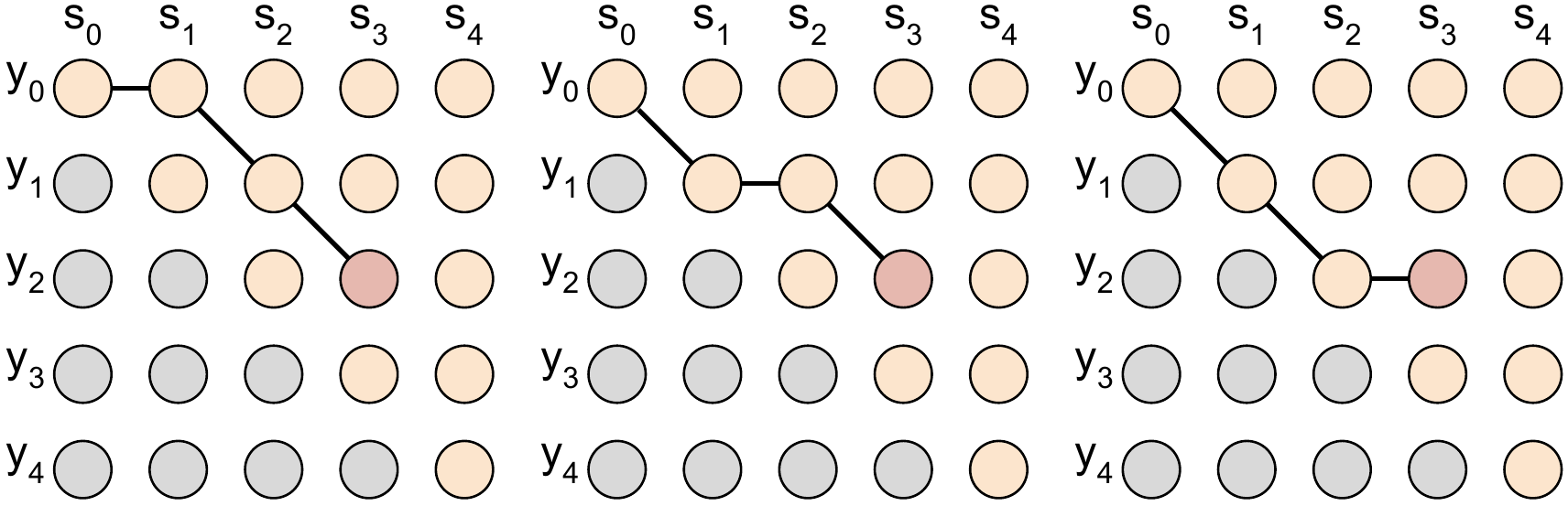}
\end{center}
\caption{Possible ways that $y_2 = s_3$: either $y_0 = s_1, y_1 = s_2, y_2 = s_3$ or $y_0 = s_0, y_1 = s_1, y_2 = s_3$ or $y_0 = s_0, y_1 = s_1, y_2 = s_3$.  \Cref{eq:p_y_m_eq_s_n} sums over these possibilities to compute $p(y_2 = s_3)$.  Gray and yellow nodes nodes indicate invalid and valid input-output pairings respectively.}
\label{fig:paths}
\end{figure}

\begin{figure}[h]
\begin{center}
\includegraphics[width=\textwidth]{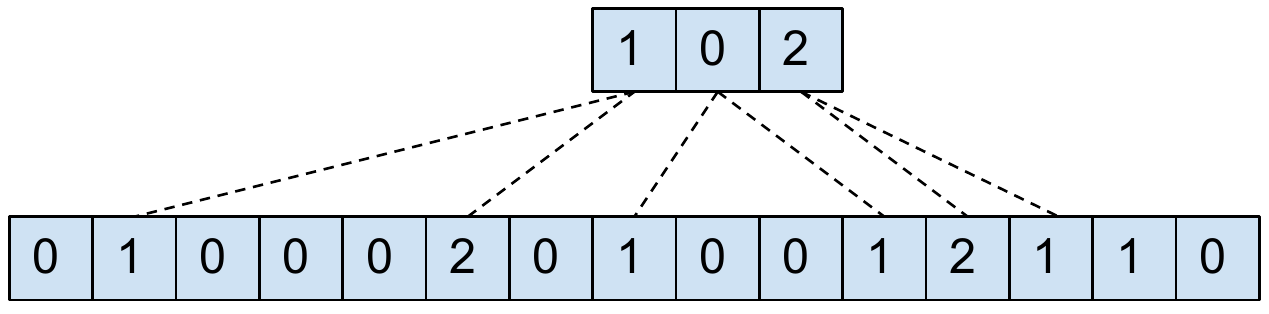}
\end{center}
\caption{Example input (bottom) and output (top) sequence for the toy problem described in \cref{sec:experiment}.  Dashed lines indicate which values in the input sequence cause each output element.}
\label{fig:toy_example}
\end{figure}

\end{document}